# False Detection (Positives and Negatives) in Object Detection

*An Experimental Study*

Subrata Goswami
*sgoswami@umich.edu*

## Abstract

Object detection is a very important function of visual perception systems. Since the early days of classical object detection based on HOG to modern deep learning based detectors, object detection has improved in accuracy. Two stage detectors usually have higher accuracy than single stage ones. Both types of detectors use some form of quantization of the search space of rectangular regions of image. There are far more of the quantized elements than true objects. The way these bounding boxes are filtered out possibly results in the false positive and false negatives. This empirical experimental study explores ways of reducing false positives and negatives with labelled data.. In the process also discovered insufficient labelling in Openimage 2019 Object Detection dataset.

## Introduction

Advances in camera, image processing, computing has made visual perception readily feasible for many applications such as robots, physical security, etc. Visual perception systems rely on one of more object detectors. Since the early days of object detectors based on HOG features to modern day deep learning based detectors such as Faster-RCNN, CornerNet, etc., object detection has improved in accuracy. See [1] for a very brief and insightful overview of object detection. A general object detector can be viewed as an image classifier with an object localizer. Image classification is a fairly mature subject with high quality networks such as ResNet. The localizer is where most of the complexities of an object detector resides and still an active area of research.

In most object detectors, localizer is based on some form of regioning of image. Because there can be almost infinite number of regions ( e.g. for a 640x480 image there can be 640x480x640x480 ~ 94 billion ), some form of quantization of region is applied. One such quantization is called Anchor Boxes, and is used in many two stage and one stage detectors. A consequence of the regioning approach is that the detector has to filter out many regions that do not contain any object. As such even with a very accurate filter, short of 100% accuracy, can result in sizeable number of false positives.

This paper goes through a number of experiments that captures false positives and negatives and looks at ways to reduce them.

## Datasets

Over the last few years a number of data sets have been developed. Some of the most well known ones are ImageNet, PASCAL-VOC, COCO, KITTI, etc [,2,3,4,5]. Most of these datasets have several types of labels - classification, detection, semantic and instance segmentation, multi-modal (e.g. LIDAR and camera) annotations etc. All of these datasets have also been vetted by many for their quality and accuracy. The following table summarizes the accuracy achieved for some of the datasets in object detection.

PASCAL-VOC is the oldest well known dataset for object detection. It has 20 classes in 11,530 images containing 27,450 annotated objects and 6,929 segmentations

One of the newest dataset is OpenImages 2019 Object Detection. This dataset is a subset of the OpenImages V5 dataset, and contains 1.7M images, 12.2M bounding boxes, and 500 categories consisting of five different levels

**Table 1:** Highest Object Detection accuracy on well known data sets [ 6, 7, 8, 9 ] .

| Dataset | Accuracy |
|---|---|
| PASCAL-VOC | 92.9(mAP) |
| COCO | 72.9 (IOU 0.5 C17) |
| KITTI | 82.5 ( pedestrian) |
| ImageNet 2019 | 68.3 |

The current highest accuracy on PASCAL-VOC is achieved by an ensemble method [6]. The base accuracy obtained by the Faster RCNN team back in 2015 was 75,5%.

## Related Works

As mentioned in the previous section, the metrics for determining accuracy is Mean Average Precision (mAP). mAP is determined by the fraction of True Positive ( TP) over the sum of TP and False Positive (FP). Hence it tends to just 0 when FP is high. However, it does not account for False Negatives (FN). FN is included in another metric called Recall. Dataset used to train and test the detector has to closely match intended inference images for mAP to remain high. If that is not the case, then mAP gradually drifts lower. One common technique to keep mAP high, is to periodically retrain detector with additional labelled data from newer images. This keeps the training data in close similarity to observed inference data.

In [10], the authors found that classification inaccuracy is a major contributor to FP. They addressed this by adding a second and deeper classifier in parallel that outputs more accurate classification.

In [11]], authors convert the detector to a class/non-class binary detector. False Positive is reduced by training on weakly labelled negative samples.

Negative examples are also used in Contrastive Learning type unsupervised methods. Where distance between positive and negative images are increased in the latent space [12].

## Measurement Metrics

Although, VOC way of measuring effectiveness is a well established method, it does not translate to detector's performance in all cases. PASCAL-VOC way of computing can result in significant over-estimation of AP. It primarily stems from the way the AP is calculated from the Precision-Recall curve. It computes a version of the measured precision/recall curve with precision monotonically decreasing, by setting the precision for recall r to the maximum precision obtained for any recall r′ ≥ r [13]. An extreme case would be where precision is 1.0 and recall 1.0, but 0 for all other recall values. PASCA-VOC method would calculate an AP of 1.0 whereas in realty the detector has a poor precision, almost 0. Another factor that may artificially increase the AP is how the TP values are distributed during the calculation of the precision-recall curve. If all the TP values are bunched at the beginning, the AP will come out significantly higher.

Hence an alternative method is used - named Global Precision and Global Recall. In this method, we gather the aggregate count of True Positive, False Positives, and False Negative per class over all images. This is much simpler to calculate and closer to how object detectors are used.

## Framework

To run the experiments, a dataset has to be created or chosen. To validate a new dataset is an expensive process. Hence the choice to use one of the well known object detection datasets. Most open source object detectors use COCO for training. That leaves KITTI, PASCAL-VOC and ImageNet, Both KITTI and PASCAL-VOC are fairly old. ImageNet is the most recent and more extensive dataset, This paper uses a subset of the ImageNet 2019 dataset.

An object detection network also has to be chosen or developed. Developing a new network from ground up is also a very expensive task. Hence the decision to use a fairly well regarded open source network called Tensorpack [14]. The Faster RCNN model pre-trained on COCO from Tensorpack was chosen. It has been used in another experimental study by the author [8] and found to be a good candidate. The choice of the same network and configuration provides a way to do relative comparison between different experiments.

As Imagenet has very large number of labelled images, it will be computationally very expensive to use the full dataset. Hence a subset is used. The subset consists of Test, Validation and Train0 data files with 21 selected classes (out of 500). A negative data set is also created that consists of images that has no human annotation of the selected classes. Table 1, 2, and 3 summarizes the selected class data for the 3 ImageNet 2019 datasets - Train0, Test and Validation. 6 different sets were created Train0-Positive/Negative, Test-Positive/Negative, Validation-Positive/Negative. The X-Negative sets contains all images that do not have any of the selected classes. Similarly X-Positive consists of images that have at least 1 object of the selected classes.

## Loss Function

The loss function used in MaskRCNN has 3 primary components - rpn_loss, head_loss (Fast-RCNN loss) and regularization. Borth rpn_loss and head_loss consists of label_loss and box_loss. The label_loss'es are Softmax Cross Entropy. Softmax Cross Entropy tends to infinity for foth perfect false positive and negative. For an illustration see the following table for Binary Cross Entropy.

**Table 2 :** Binary Cross Entropy Loss for false positives and negatives. Loss = -p.log(ph) - (1-p).log(1-ph) .

| Label Probability | Prediction probability | Loss | Note |
|---|---|---|---|
| 0 | 0 | 0 | True -ve |
| 1 | 0 | infinity | False -ve |
| 0 | 1 | infinity | False +ve |

| 1 | 1 | 0 | True +ve |

The way the box_loss is computed, it is 0 if there is no ground truth boxes. This results from the fact that in Tensorpack, all proposal boxes have to have >0 IOU with ground truth boxes to be propagated to output of Fast RCNN head, and hence to be included in the loss calculation.

There is also a difference between the loss value used in backpropagation versus the one displayed. The displayed values are initialized at non-0 values and updated with a moving average algorithm according to a decay factor. This is not pertinent to weight update, but can be cause of confusion.

## Annotation Quality

Because of the volume of data in ImageNet, it was not possible to verify each image and annotations manually. A few anecdotal examples of incorrect or incomplete annotations are provided in the Appendix.

## Experiments

The following are the experiments conducted on the ImageNet 2019 data subsets. In all the experiments, the pretrained checkpoint provided by Tensorpack for Resnet101 is used, unless otherwise mentioned.

1. Training on the Train0 full data set, and validating on the Validation data set for 7 epochs.
2. Training on the Train0 positive data set, and validating on the Validation data set for 8 epochs..
3. Training on the Train0 negative data set with the checkpoint from 2 above and validating on the Validation data set.
4. Training on the Train0 negative set, and validating on the Validation data set.
5. Train as in 3 above, but with short epoch step size of 100 images.
6. Training with a synthetic uniform gray scale image and validating on the Validation data set. Each spoch was 100 samples of the same gray image.

## Results and Discussions

The results of the experiments 1 through 4 are detailed in Table 4. As can be seen there is improvement in precision in Experiment 1 versus Experiment 2 for all classes except for 2. However, recall results were mixed, some improvement in 1 vs experiment 2 and vice-versa.

Experiments 3 and 4 did not produce any results, hence the corresponding columns are left blank. This probably happened because the network becomes degenerated from confusing (non-comprehensive and incorrect ) labels. Also looked at the checkpoints tensor values generated by Experiments 1 and 3. Part of values are listed in List 1 and List 2 for Experiment 1 and Experiment 3 respectively. In general the most apparent difference is in gradients, which in Experiment 3 are all 0, which would imply no further changes in the weight for iterations. Which also implies that loss has become 0.

To look at how label quality impacts training with negative samples, 2 additional experiments were conducted, - Experiments 5 and 6. The results from Experiments 5 and 6 are shown in Table 5 and Table 6 respectively. In Table 6 the Precision and Recall numbers remained relatively stable. However, in Table 5 they rapidly deteriorated towards 0. One of the reasons for this deterioration towards 0, could be due to incorrectness and incompleteness in annotations If the gray image in Experiment 6 is replaced with a uniform random RGB image, similar results were obtained.

## Conclusion

A number of experiments were conducted on Imagenet 2019 Object Detection dataset with a goal to address false detection issues. Slight improvements were observed when positive and negative images are used together in training. It was observed that when negative images alone are use to fine tune, the network rapilly degenerates, most likely due to incomplete labelling. The same issue was not observed in fine tunning on a grayscale image.

Further work with complete and accurately labelled images can be used to confirm the above findings.

## References

1. Ross Girshick, Object Detection as a Machine Learning Problem https://www.youtube.com/watch?v=her4_rzx09o
2. Open Images 2019 - Object Detection - Detect objects in varied and complex image, https://www.kaggle.com/c/open-images-2019-object-detection
3. Everingham, M., Eslami, S. M. A., Van Gool, L., Williams, C. K. I., Winn, J. and Zisserman, A. , The PASCAL Visual Object Classes Challenge: A Retrospective,International Journal of Computer Vision, 111(1), 98-136, 2015 .
4. Lin TY, Maire M, Belongie S, Hays J, Perona P, Ramanan D, Doll´ar P, Zitnick CL (2014) Microsoft coco: Common objects in context. In: Proc. of the European Conf. on Computer Vision (ECCV)
5. A. Geiger, P. Lenz, C. Stiller, and R. Urtasun. Vision meets robotics: The kitti dataset. International Journal of Robotics Research (IJRR), 2013
6. PASCAL-VOC object detection leaderboard http://host.robots.ox.ac.uk:8080/leaderboard/displaylb.php?challengeid=11&compid=4
7. KITTI object detection leaderboard , http://www.cvlibs.net/datasets/kitti/eval_object.php
8. COCO object detection leader board http://cocodataset.org/#detection-eval
9. Open Images 2019 - Object Detection, https://www.kaggle.com/c/open-images-2019-object-detection/leaderboard
10. Decoupled Classification Refinement: Hard False Positive Suppression for Object Detection, https://arxiv.org/pdf/1810.04002.pdf
11. Incorporating Negative Sample Training for Ship Detection Based on Deep Learning, , https://www.ncbi.nlm.nih.gov/pmc/articles/PMC6387301/
12. Bootstrap Your Own Latent A New Approach to Self-Supervised Learning, https://arxiv.org/pdf/2006.07733.pdf .
13. The PASCAL Visual Object Classes Challenge, 2010 (VOC2010) Development Kit, http://host.robots.ox.ac.uk/pascal/VOC/voc2010/devkit_doc_08-May-2010.pdf
14. Tensorpack, https://github.com/tensorpack/tensorpack

## Tables and Figures

**Table 1:** Distribution of selected class objects and images in the ImageNet 2019 Train0 set.

| | bbox | neg images | pos images | pos % |
|---|---|---|---|---|
| total images | | | 156541 | |
| neg images | | | 79112 | |
| pos images | | | 77429 | |
| Airplane | 2154 | 155360 | 1181 | 0.75 |
| Car | 16955 | 150523 | 6018 | 3.84 |
| Dog | 2410 | 154847 | 1694 | 1.08 |
| Footwear | 49171 | 147858 | 8683 | 5.55 |
| Human arm | 15634 | 153263 | 3278 | 2.09 |
| Human body | 13626 | 152780 | 3761 | 2.40 |
| Human ear | 1689 | 155664 | 877 | 0.56 |
| Human eye | 6312 | 154511 | 2030 | 1.30 |
| Human face | 66174 | 135139 | 21402 | 13.67 |
| Human foot | 807 | 156197 | 344 | 0.22 |
| Human hand | 6111 | 154580 | 1961 | 1.25 |
| Human head | 15322 | 152811 | 3730 | 2.38 |
| Human leg | 6178 | 154897 | 1644 | 1.05 |
| Human mouth | 3562 | 154512 | 2029 | 1.30 |
| Human nose | 4517 | 154286 | 2255 | 1.44 |
| Man | 92320 | 132122 | 24419 | 15.60 |
| Person | 67540 | 140601 | 15940 | 10.18 |
| Tire | 8697 | 154291 | 2250 | 1.44 |
| Tree | 69130 | 137439 | 19102 | 12.20 |
| Vehicle registration plate | 869 | 155927 | 614 | 0.39 |
| Woman | 49524 | 139012 | 17529 | 11.20 |

**Table 2:** Distribution of selected class objects and images in the ImageNet 2019 Test set.

| total images | | | 112194 | |
|---|---|---|---|---|
| neg images | | | 50197 | |
| pos images | | | 61997 | |
| | bbox | neg image | pos image | pos % |
| Airplane | 3306 | 109738 | 2456 | 3.96 |
| Car | 30153 | 96906 | 15288 | 24.66 |
| Dog | 5856 | 107359 | 4835 | 7.80 |
| Footwear | 26882 | 106388 | 5806 | 9.36 |
| Human arm | 34067 | 101758 | 10436 | 16.83 |
| Human body | 34311 | 95291 | 16903 | 27.26 |
| Human ear | 3304 | 110219 | 1975 | 3.19 |
| Human eye | 16257 | 106266 | 5928 | 9.56 |
| Human face | 17008 | 102902 | 9292 | 14.99 |
| Human foot | 1576 | 111537 | 657 | 1.06 |
| Human hand | 15185 | 105920 | 6274 | 10.12 |
| Human head | 27898 | 100941 | 11253 | 18.15 |
| Human leg | 18818 | 106695 | 5499 | 8.87 |
| Human mouth | 8161 | 106254 | 5940 | 9.58 |
| Human nose | 13219 | 104419 | 7775 | 12.54 |
| Man | 26231 | 100844 | 11350 | 18.31 |
| Person | 53385 | 90506 | 21688 | 34.98 |
| Tire | 21183 | 105466 | 6728 | 10.85 |
| Tree | 26477 | 101040 | 11154 | 17.99 |
| Vehicle registration plate | 2843 | 110129 | 2065 | 3.33 |
| Woman | 14551 | 103935 | 8259 | 13.32 |

**Table 3:** Distribution of selected class objects and images in the ImageNet 2019 Valdation set.

| | bbox | neg image | pos image | pos % |
|---|---|---|---|---|
| total images | | | 37306 | |
| neg images | | | 20448 | |
| pos images | | | 16858 | |
| Airplane | 1032 | 36548 | 758 | 2.03 |
| Car | 9924 | 32211 | 5095 | 13.66 |
| Dog | 1937 | 35720 | 1586 | 4.25 |
| Footwear | 9110 | 35399 | 1907 | 5.11 |
| Human arm | 11711 | 33774 | 3532 | 9.47 |
| Human body | 11744 | 31620 | 5686 | 15.24 |
| Human ear | 1147 | 36623 | 683 | 1.83 |
| Human eye | 5450 | 35344 | 1962 | 5.26 |
| Human face | 5594 | 34182 | 3124 | 8.37 |
| Human foot | 482 | 37107 | 199 | 0.53 |
| Human hand | 5031 | 35250 | 2056 | 5.51 |
| Human head | 8765 | 33595 | 3711 | 9.95 |
| Human leg | 5818 | 35507 | 1799 | 4.82 |
| Human mouth | 2752 | 35327 | 1979 | 5.30 |
| Human nose | 4499 | 34694 | 2612 | 7.00 |
| Man | 8493 | 33572 | 3734 | 10.01 |
| Person | 16753 | 30162 | 7144 | 19.15 |
| Tire | 6742 | 35079 | 2227 | 5.97 |
| Tree | 8077 | 33695 | 3611 | 9.68 |
| Vehicle registration plate | 987 | 36582 | 724 | 1.94 |
| Woman | 4767 | 34531 | 2775 | 7.44 |

Goswami et al. |Impact of Data Quality on Deep Neural Network Training - An Experimental Study**Table 4: Summary results, precision and recall, from Experiments 1 through 4. Note that Experiments 3 and 4 did not produce any results, probably because the network became degenerated from confusing (non-comprehensize and incorrect) labels. There is improvement in experiment 1 where all positive and negative images were included vs Experiment 2 where only positives images were included.**

| class name | 1 - all train0 precision | recall | 2 pos train0 precision | recall | 3 and 4 - neg train0 precision | recall |
|---|---|---|---|---|---|---|
| Airplane | 16.85 | 92.01 | 12.83 | 89.23 | | |
| Car | 25.73 | 90.82 | 18.63 | 90.8 | | |
| Dog | 45.81 | 92.38 | 20.5 | 93.34 | | |
| Footwear | 18.05 | 67.54 | 15.07 | 64.98 | | |
| HumanArm | 31.31 | 39.36 | 20.38 | 48.6 | | |
| HumanBody | 51.78 | 30.47 | 50.17 | 32.57 | | |
| HumanEar | 35.82 | 37.23 | 28.64 | 40.98 | | |
| HumanEye | 48.89 | 49.82 | 24.15 | 43.39 | | |
| HumanFace | 16.54 | 86.58 | 13.3 | 85.5 | | |
| HumanFoot | 15.56 | 38.05 | 5.14 | 41.29 | | |
| HumanHand | 34.22 | 37.14 | 15.32 | 40.75 | | |
| HumanHead | 28.2 | 64.32 | 22.28 | 63.64 | | |
| HumanLeg | 24.38 | 37.76 | 15.56 | 46.96 | | |
| HumanMouth | 45.6 | 47.53 | 23.4 | 49.93 | | |
| HumanNose | 59.21 | 42.15 | 38.69 | 38.48 | | |
| Man | 13.34 | 88.87 | 10.07 | 89.29 | | |
| Person | 14.63 | 71.24 | 9.55 | 73.9 | | |
| Tire | 21.83 | 82.65 | 20.84 | 80.26 | | |
| Tree | 8.67 | 73.99 | 5.34 | 76.96 | | |
| VehicleRegistrationPlate | 18.99 | 69.3 | 21.19 | 62.41 | | |
| Woman | 14.3 | 84.91 | 15.82 | 78.79 | | |

**Table 5: Summary results, precision and recall, from Experiment 5 over 300 iterations of training.**

| Iteration | 100 | | 200 | | 300 | |
|---|---|---|---|---|---|---|
| class name | Precision | Recall | Preision | Recall | Precision | Recall |
| Airplane | 64.09 | 59 | 0 | 0 | 0 | 0 |
| Car | 54.67 | 71.41 | 56.68 | 7.34 | 25 | 2.2 |
| Dog | 70.62 | 50.3 | 0 | 0 | 0 | 0 |
| Footwear | 25.95 | 51.8 | 51.44 | 14.24 | 60 | 6.04 |
| HumanArm | 26.35 | 12.34 | 28.57 | 0.07 | 0 | 0 |
| HumanBody | 48.83 | 10.11 | 0 | 0 | 0 | 0 |
| HumanEar | 47.13 | 7.44 | 0 | 0 | 0 | 0 |
| HumanEye | 59.43 | 34.46 | 78.26 | 1.43 | 0 | 0 |
| HumanFace | 21.25 | 77.55 | 26.05 | 52.81 | 23.08 | 22.15 |
| HumanFoot | 20.93 | 9.86 | 0 | 0 | 0 | 0 |
| HumanHand | 39.1 | 12.26 | 0 | 0 | 0 | 0 |
| HumanHead | 48.76 | 29.76 | 69.23 | 0.22 | 0 | 0 |
| HumanLeg | 25.34 | 9.1 | 0 | 0 | 0 | 0 |
| HumanMouth | 49.65 | 41.06 | 38.89 | 0.64 | 0 | 0 |
| HumanNose | 66.32 | 28.16 | 84.62 | 7.66 | 0 | 0 |
| Man | 17.49 | 69.53 | 16.02 | 6.95 | 15.79 | 1.6 |
| Person | 18.31 | 45.08 | 24.21 | 6.57 | 30 | 1.17 |
| Tire | 33.12 | 63.75 | 35.37 | 18.79 | 71.43 | 8.06 |
| Tree | 15.56 | 42.74 | 25.62 | 3.32 | 66.67 | 1.94 |
| VehicleRegistrationPlate | 28.68 | 28.18 | 0 | 0 | 0 | 0 |
| Woman | 14.06 | 79.38 | 20.48 | 1.57 | 50 | 0.28 |

**Table 6: Summary results, precision and recall, from Experiment 6 over 300 iterations of training.**

|  | 100 |  | 200 |  | 300 |  |
| --- | --- | --- | --- | --- | --- | --- |
| class_name | Precision | Recall | Preision | Recall | Precision | Recall |
| Airplane | 9.28 | 90.79 | 8.7 | 90.88 | 8.59 | 90.79 |
| Car | 14.08 | 91.78 | 13.13 | 91.79 | 12.97 | 91.73 |
| Dog | 15.16 | 94.17 | 14.49 | 93.96 | 14.31 | 94.11 |
| Footwear | 7.28 | 70.42 | 5.88 | 70.66 | 5.57 | 70.93 |
| HumanArm | 15.43 | 51.72 | 14.29 | 51.27 | 13.97 | 50.95 |
| HumanBody | 46.61 | 36.11 | 45.55 | 35.86 | 45.44 | 35.78 |
| HumanEar | 21.13 | 46.12 | 19.15 | 47.25 | 18.63 | 47.69 |
| HumanEye | 14.42 | 52.28 | 12.4 | 53.72 | 11.79 | 54.11 |
| HumanFace | 9.98 | 87.02 | 9.08 | 87 | 8.83 | 87.02 |
| HumanFoot | 4.77 | 43.15 | 4.57 | 41.29 | 4.55 | 40.87 |
| HumanHand | 13.42 | 46.63 | 12.92 | 47.29 | 12.79 | 47.33 |
| HumanHead | 17.96 | 65.49 | 16.7 | 65.41 | 16.39 | 65.35 |
| HumanLeg | 12.43 | 48.4 | 11.9 | 47.27 | 11.84 | 47.1 |
| HumanMouth | 8.79 | 61.19 | 6.5 | 62.32 | 5.93 | 62.74 |
| HumanNose | 22.53 | 49.26 | 19.18 | 51.14 | 18.26 | 51.67 |
| Man | 8.47 | 89.01 | 8.16 | 88.78 | 8.1 | 88.64 |
| Person | 9.09 | 73.37 | 8.97 | 72.68 | 8.97 | 72.45 |
| Tire | 15.94 | 82.05 | 14.87 | 81.76 | 14.67 | 81.71 |
| Tree | 4.32 | 77.34 | 4.08 | 76.81 | 4.03 | 76.55 |
| VehicleRegistrationPlate | 15.57 | 67.38 | 14.17 | 67.48 | 13.81 | 67.48 |
| Woman | 14.06 | 79.38 | 13.87 | 78.81 | 13.87 | 78.79 |

**List 1: Snippet of the checkpoint fastrcnn/class tensor from Experiment 1**

```
tensor_name:  fastrcnn/class/W
[[-1.8081831e-02  2.1900643e-02  5.1620705e-03 ... -2.8967526e-02
  -1.3970777e-02 -1.1211102e-02]
 [-3.5974316e-02 -5.6404318e-03  6.1590788e-03 ...  3.5809998e-03
   7.1761970e-05 -4.3327091e-03]
 [ 1.1485914e-02 -8.1279501e-03 -4.1792294e-04 ...  3.3695920e-04
   1.3076856e-03 -3.0689861e-03]
 ...
 [-4.6240162e-02  3.8433021e-03 -6.3901390e-03 ... -3.6396470e-04
   1.2519509e-03 -5.3341035e-03]
 [-8.2251383e-03 -2.1703601e-02 -5.8959313e-03 ...  1.1010152e-02
   1.8961253e-02  1.3649580e-03]
 [-3.4825861e-02 -7.8396661e-05 -1.0385877e-02 ... -7.9760365e-03
  -1.1337606e-02  2.8376965e-02]]
tensor_name:  fastrcnn/class/W/AccumGrad
[[ 2.0336017e-03  3.3454483e-04 -2.1219619e-03 ... -8.3851410e-06
  -4.0954023e-06 -3.2174514e-06]
 [ 1.2843193e-03  1.8989190e-04 -4.6472540e-04 ...  1.9690116e-05
   1.1403985e-07 -1.0808437e-06]
 [ 2.4312795e-03  6.9452079e-05 -2.6685206e-04 ...  3.5214616e-05
   4.9639226e-07 -8.3038572e-07]
 ...
 [ 6.4088672e-02  2.5183507e-04 -1.8526094e-03 ...  5.2279538e-05
   1.1937734e-06  1.4617215e-06]
 [ 4.0844943e-02  1.2358313e-03 -8.2572503e-03 ...  1.7146411e-04
   6.7747942e-06  1.5695069e-06]
 [ 4.4862190e-03  1.0869551e-03 -1.0982313e-02 ...  2.3378079e-05
  -3.0503038e-06  9.2045029e-06]]
tensor_name:  fastrcnn/class/W/Momentum
[[ 3.38412970e-02 -8.30419585e-02 -7.64503889e-03 ... -8.88858282e-04
  -8.62269881e-05  3.36249359e-04]
 [ 1.27878943e-02 -2.93821674e-02 -1.30038254e-03 ... -4.72720060e-03
  -7.14763810e-05  2.51758611e-04]
 [-7.97510054e-03 -6.29643630e-03 -9.32441850e-04 ... -1.28085539e-03
  -1.49912275e-05  9.11218885e-05]
 ...
 [-2.03980282e-02 -5.77657891e-04 -4.73211985e-03 ... -4.21997812e-03
  -1.24974526e-04  9.07886715e-04]
 [ 1.59230698e-02 -1.83396041e-02 -1.65999879e-03 ... -1.74125482e-03
  -7.81440176e-03  1.00140192e-03]
 [ 4.51412834e-02 -1.40836695e-02 -5.75931789e-03 ... -6.31267962e-04
  -5.12379411e-05  1.79236836e-03]]
tensor_name:  fastrcnn/class/b
[ 1.0415046   0.14411277 -0.03873769 -0.00517399  0.2591181  -0.1125795
  0.08348841 -0.17880213  0.14731742  0.12907633 -0.14202012 -0.08457479
```

```
 -0.15513058 -0.14866912 -0.14775841  0.0183127  -0.14224118 -0.10526124
 -0.14523657 -0.12238844 -0.14600106 -0.14835985]
tensor_name:  fastrcnn/class/b/AccumGrad
[ 1.9724005e-01  9.0487367e-03 -5.9578754e-02  3.9135310e-05
  3.0248398e-02 -3.2242198e-02 -7.8078434e-02  2.8362093e-03
 -4.4680662e-03  4.3539563e-03 -2.2194916e-02  2.0111325e-05
  3.3332184e-05 -8.3591221e-03 -2.7232917e-02  1.3869570e-04
  2.1198504e-04  1.2921541e-05 -1.2317867e-02  2.7276610e-04
  6.3011330e-06  9.6947124e-06]
tensor_name:  fastrcnn/class/b/Momentum
[-0.24559109 -0.01191232 -0.04899206  0.0568433   0.01894237  0.02644759
 -0.05353307 -0.05827032 -0.01332948  0.26956615 -0.00107717  0.04159928
  0.00897772 -0.00165643 -0.03231319  0.07338736  0.01589685 -0.00507927
 -0.01571575 -0.01928952 -0.01033568  0.00543516]
```

**List 2: Snippt of the checkpoint fastrcnn/class tensor from Experiment 3**

```
tensor_name:  fastrcnn/class/W
[[ 0.00884428 -0.00349497 -0.00136673 ... -0.00190676 -0.00624543
   0.00191707]
 [ 0.00309437  0.00128538 -0.00568914 ...  0.00227561 -0.00566468
  -0.00596222]
 [ 0.00304964 -0.00064118 -0.00014935 ...  0.00325428  0.0032456
   0.0017629 ]
 ...
 [ 0.00591924  0.00539389  0.00568572 ...  0.001324    0.00649751
  -0.00280084]
 [ 0.01719209 -0.00563596  0.00169213 ... -0.00360665 -0.00335684
  -0.00396957]
 [ 0.00328937 -0.00050531 -0.00327482 ...  0.00084169  0.00358218
  -0.00323338]]
tensor_name:  fastrcnn/class/W/AccumGrad
[[0. 0. 0. ... 0. 0. 0.]
 [0. 0. 0. ... 0. 0. 0.]
 [0. 0. 0. ... 0. 0. 0.]
 ...
 [0. 0. 0. ... 0. 0. 0.]
 [0. 0. 0. ... 0. 0. 0.]
 [0. 0. 0. ... 0. 0. 0.]]
tensor_name:  fastrcnn/class/W/Momentum
[[ 7.40042815e-05 -2.92440782e-05 -1.14360992e-05 ... -1.59547835e-05
  -5.22584414e-05  1.60410418e-05]
 [ 2.58920154e-05  1.07553624e-05 -4.76037458e-05 ...  1.90410774e-05
  -4.73990622e-05 -4.98887021e-05]
 [ 2.55177292e-05 -5.36501966e-06 -1.24969256e-06 ...  2.72300676e-05
   2.71574863e-05  1.47510345e-05]
 ...
 [ 4.95290624e-05  4.51332198e-05  4.75751003e-05 ...  1.10785722e-05
```

```
    5.43677015e-05 -2.34359431e-05]
 [ 1.43778059e-04 -4.71503081e-05  1.41678429e-05 ... -3.01703403e-05
  -2.80783806e-05 -3.32052259e-05]
 [ 2.75236707e-05 -4.22813810e-06 -2.74019767e-05 ...  7.04282229e-06
   2.99738131e-05 -2.70552173e-05]]
tensor_name:  fastrcnn/class/b
[ 0.09621707 -0.00551699 -0.00389974 -0.0039435  -0.00429787 -0.00360999
 -0.00553105 -0.00456049 -0.00413389 -0.00438579 -0.00496833 -0.0048181
 -0.00384614 -0.0048066  -0.00500841 -0.00408059 -0.00447912 -0.00550086
 -0.00497839 -0.00413672 -0.00473903 -0.00497544]
tensor_name:  fastrcnn/class/b/AccumGrad
[0. 0. 0. 0. 0. 0. 0. 0. 0. 0. 0. 0. 0. 0. 0. 0. 0. 0. 0. 0. 0. 0.]
tensor_name:  fastrcnn/class/b/Momentum
[-4.9089556e-07  4.4799012e-08  4.8186859e-08  3.4858743e-08
  4.2956728e-08  3.3210167e-08  4.3076316e-08  3.8288153e-08
  3.6907469e-08  4.0693934e-08  3.7066386e-08  4.9777896e-08
  3.7570061e-08  4.7526047e-08  3.3975152e-08  3.5122181e-08
  3.3497834e-08  4.2041901e-08  3.7695074e-08  4.3235154e-08
  5.2195848e-08  5.2692833e-08]
```

## Appendix : Annotation Issues in Imagenet

The following image, 0c33437a18e6e9f9, only has annotation for Gas stove, /m/02wv84t, but not for Man, Human Hand, human Arm, etc.

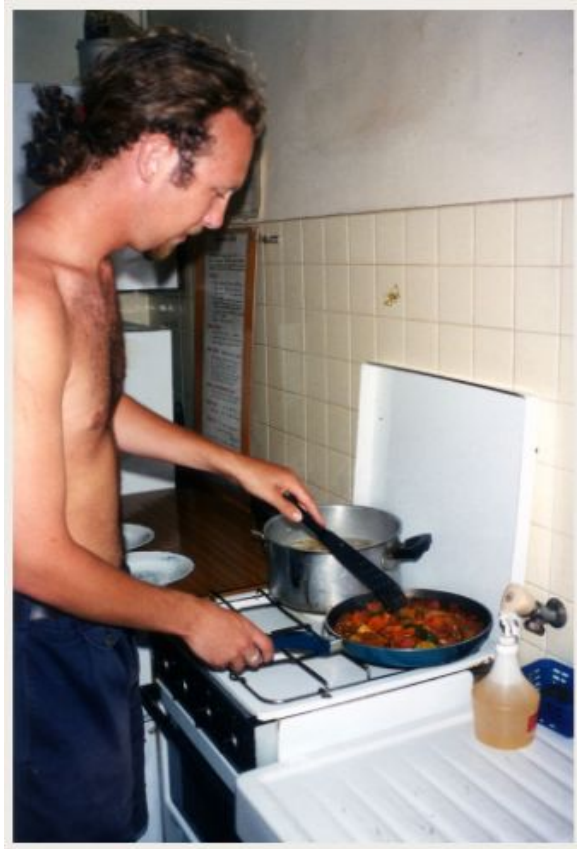

The following image, 0001eeaf4aed83f9, only has annotation for 1 Airplane, /m/0cl4p , but parts of 4 more are visible .

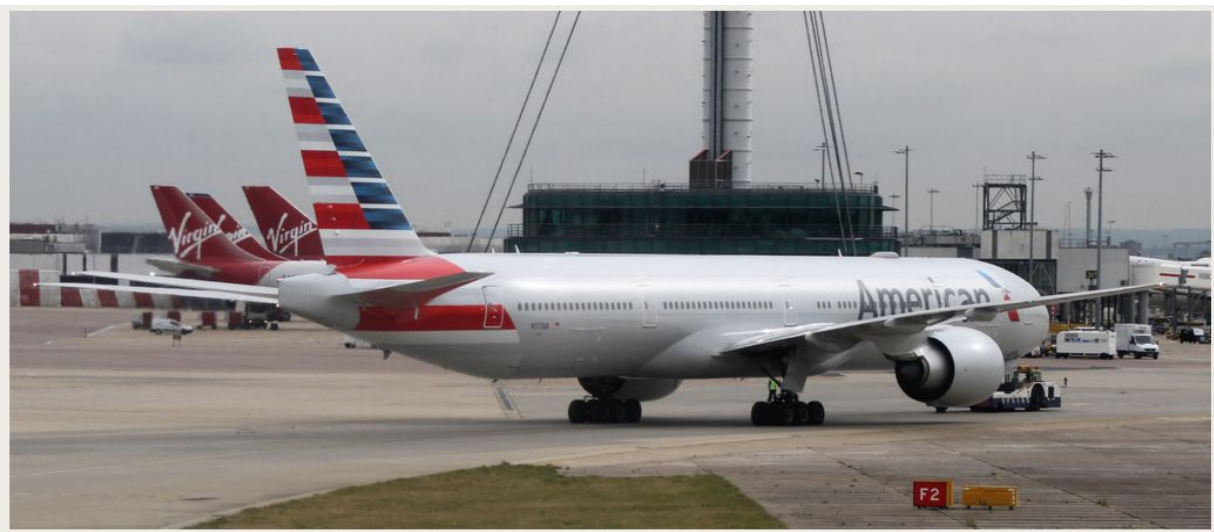

## Appendix : Annotation Issues in Imagenet